\pdfoutput=1

\documentclass[11pt]{article}

\usepackage[]{ACL2023}

\usepackage{times}
\usepackage{latexsym}
\usepackage{booktabs}
\usepackage{graphicx} 
\usepackage[normalem]{ulem}
\useunder{\uline}{\ul}{}
\usepackage{enumitem}
\usepackage[T1]{fontenc}
\usepackage{multirow}
\usepackage{changes}
\usepackage[utf8]{inputenc}

\usepackage{microtype}

\usepackage{inconsolata}
\usepackage{xspace}

\newif\ifhidecomments
\hidecommentsfalse

\newcommand{\figref}[1]{Fig.~\ref{#1}}
\newcommand{\tableref}[1]{Table~\ref{#1}}
\newcommand{\dataset}{\textsc{CHIME}}
\newcommand{\para}[1]{\noindent\textbf{#1}\xspace}
\makeatletter
\def\blfootnote{\xdef\@thefnmark{}\@footnotetext}
\makeatother

\title{\dataset{}: LLM-Assisted Hierarchical Organization of Scientific Studies for Literature Review Support}

\author{Chao-Chun Hsu$^{1\spadesuit}$, Erin Bransom$^2$, Jenna Sparks$^2$, Bailey Kuehl$^2$, \\
{\bf Chenhao Tan}$^1$, {\bf David Wadden}$^2$, {\bf Lucy Lu Wang}$^{2,3}$, {\bf Aakanksha Naik}$^2$ \\
$^1$University of Chicago, $^2$Allen Institute for AI, $^3$University of Washington \\
}

\begin{document}
\maketitle
\begin{abstract}

Literature review requires researchers to synthesize a large amount of information and is increasingly challenging as the scientific literature expands. In this work, we investigate the potential of LLMs for producing hierarchical organizations of scientific studies to assist researchers with literature review. We define hierarchical organizations as tree structures where nodes refer to topical categories and every node is linked to the studies assigned to that category. Our naive LLM-based pipeline for hierarchy generation from a set of studies produces promising yet imperfect hierarchies, motivating us to collect \dataset{}, an expert-curated dataset for this task focused on biomedicine. Given the challenging and time-consuming nature of building hierarchies from scratch, we use a human-in-the-loop process in which experts \emph{correct errors} (both links between categories and study assignment) in LLM-generated hierarchies. \dataset{} contains 2,174 LLM-generated hierarchies covering 472 topics, and expert-corrected hierarchies for a subset of 100 topics. Expert corrections allow us to quantify LLM performance, and we find that while they are quite good at generating and organizing categories, their assignment of studies to categories could be improved. 
We attempt to train a corrector model with human feedback which improves study assignment by $12.6$ F1 points. We release our dataset and models to encourage research on developing better assistive tools for literature review.\footnote{The \dataset~dataset and models are available at \url{https://github.com/allenai/chime}.}

\end{abstract}
\blfootnote{
\hspace{-19pt}
$^\dagger$ Correspondence to \texttt{aakankshan@allenai.org}. \\
$^\spadesuit$Work done as an intern at Allen Institute for AI.
}

\section{Introduction}

\begin{figure}[!h]
    \centering
    \includegraphics[width=0.9\columnwidth]{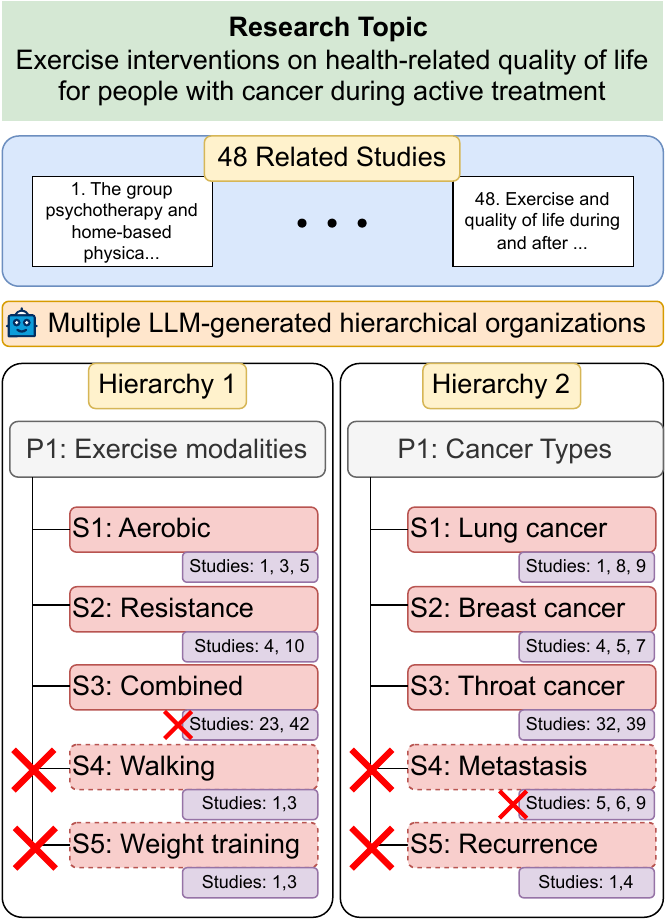}
    \caption{Given a set of related studies on a topic, we use LLMs to identify top-level categories focusing on different views of the data (such as P1 and P2), generate multiple hierarchical organizations, and assign studies to different categories. However, these categories and study assignments can contain errors. As illustrated in the figure, the categories \texttt{Walking} and \texttt{Weight training} are not coherent with their siblings ($S1-S3$) in hierarchy 1 since they are more specific, and the categories \texttt{Metastasis} and \texttt{Recurrence} are incorrectly assigned to the parent category in hierarchy 2 since they are not types of cancer.
    }
    \label{fig:example}
\end{figure}

Literature review, the process by which researchers synthesize many related scientific studies into a higher-level organization, is valuable but extremely time-consuming. For instance, in medicine, completing a review from registration to publication takes 67 weeks on average \cite{borah2017analysis} and given the rapid pace of scholarly publication, reviews tend to go out-of-date quickly \cite{shojania2007quickly}. This has prompted development of tools for efficient literature review \cite{altmami2022automatic}. Most tools have focused on automating review generation, treating it as a multi-document summarization task \cite{mohammad-etal-2009-using,jha-etal-2015-content,Wallace2020GeneratingN,deyoung-etal-2021-ms,liu2023generating}, sometimes using intermediate structures such as hierarchies/outlines to better scaffold generation \cite{zhu-etal-2023-hierarchical}, with limited success. However, recent work on assessing the utility of NLP tools like LLMs for systematic review reveals that domain experts prefer literature review tools to be assistive instead of automatic \cite{yun-etal-2023-appraising}.

Motivated by this finding, we take a different approach and focus on the task of generating hierarchical organizations of scientific studies to assist literature review. As shown in Figure~\ref{fig:example}, a hierarchical organization is a tree structure in which nodes represent topical categories and every node is linked to a list of studies assigned to that category. Inspired by the adoption of LLMs for information organization uses such as clustering \cite{viswanathan2023large} and topic modeling \cite{Pham2023TopicGPTAP}, we investigate the potential of generating hierarchies with a naive LLM-based approach, and observe that models produce promising yet imperfect hierarchies out-of-the-box.

To further assess and improve LLM performance, we collect \dataset{} (\textbf{C}onstructing \textbf{HI}erarchies of bio\textbf{M}edical \textbf{E}vidence), an expert-curated dataset for hierarchy generation. Since building such hierarchies from scratch is very challenging and time-consuming, we develop a human-in-the-loop protocol in which experts \emph{correct errors} in preliminary LLM-generated hierarchies. During a three-step error correction process, experts assess the correctness of both links between categories as well as assignment of studies to categories, as demonstrated in Figure~\ref{fig:example}. Our final dataset consists of two subsets: (i) a set of 472 research topics with up to five LLM-generated hierarchies per topic (2,174 total hierarchies), and (ii) a subset of 100 research topics sampled from the previous set, with 320 expert-corrected hierarchies. 
 
Expert-corrected hierarchies allow us to better quantify LLM performance on hierarchy generation. We observe that LLMs are already quite good at generating and organizing categories, achieving near-perfect performance on parent-child category linking and a precision of $77.3\%$ on producing coherent groups of sibling categories. However, their performance on assigning studies to relevant categories ($61.5\%$ F1) leaves room for improvement. We study the potential of using \dataset{} to train ``corrector'' agents which can provide feedback to our LLM-based pipeline to improve hierarchy quality. Our results show that finetuning a FLAN-T5-based corrector and applying it to LLM-generated hierarchies improves study assignment by $12.6$ F1 points. We release our dataset containing both LLM-generated and expert-corrected hierarchies, as well as our LLM-based hierarchy generation and correction pipelines, to encourage further research on better assistive tools for literature review. 

In summary, our key contributions include:
\begin{itemize}[leftmargin=*,topsep=0pt]
\setlength\itemsep{-0.5em}
\item We develop an LLM-based pipeline to organize a collection of papers on a research topic into a labeled, human-navigable concept hierarchy.
\item We release \dataset{}, a dataset of 2174 hierarchies constructed using our pipeline, including a ``gold'' subset of 320 hierarchies checked and corrected by human experts.
\item We train corrector models using \dataset{} to automatically fix errors in LLM-generated hierarchies, improving accuracy of study categorization by 12.6 F1 points.
\end{itemize}

\section{Generating Preliminary Hierarchies using LLMs}
The first phase of our dataset creation process focuses on using LLMs to generate preliminary hierarchies from a set of related studies, which can then be corrected by experts. We describe our process for collecting sets of related studies and our LLM-based hierarchy generation pipeline. 

\subsection{Sourcing Related Studies}
We leverage the Cochrane Database of Systematic Reviews\footnote{\url{https://www.cochranelibrary.com/cdsr/reviews}} to obtain sets of related studies, since the systematic review writing process requires experts to extensively search for and curate studies relevant to review topics. We obtain all systematic reviews and the corresponding studies included in each review from the Cochrane website \cite{Wallace2020GeneratingN}. We then filter this set of systematic reviews to only retain those including at least 15 and no more than 50 corresponding studies. We discard reviews with very few studies since a hierarchical organization is unlikely to provide much utility, while reviews with more than 50 studies are discarded due to the inability of LLMs to effectively handle such long inputs \cite{liu2023lost}. Our filtering criteria leave us with 472 systematic reviews (or sets), each including an average of 24.7 studies, which serve as input to our hierarchy generation pipeline.

\subsection{Hierarchy Generation Pipeline}
\begin{figure}[t!]
    \centering
    \includegraphics[width=.95\columnwidth]{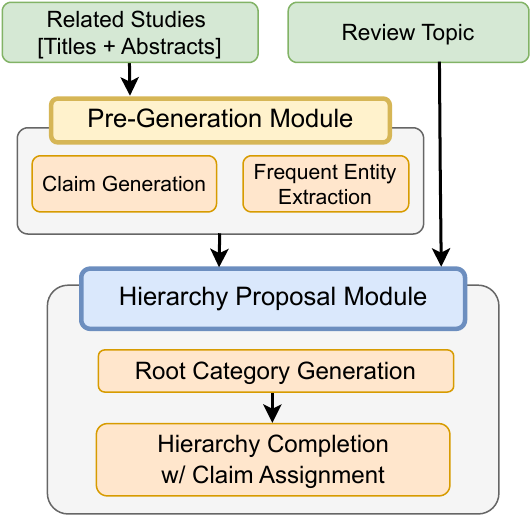}
    \caption{LLM-based pipeline for preliminary hierarchy generation given a set of related studies on a topic.
    }
    \label{fig:pipeline}
\end{figure}

Prior work on using LLMs for complex tasks has shown that decomposing the task into a series of steps or sub-tasks often elicits more accurate responses \cite{kojima2022large,wei2022chain,d2024marg}. Motivated by this, we decompose hierarchy generation from a set of related scientific studies into three sub-tasks: (i) compressing study findings into concise claims, (ii) initiating hierarchy generation by generating \textit{root} categories, and (iii) completing hierarchy generation by producing remaining categories and organizing claims under them. Our hierarchy generation pipeline consists of a \textit{pre-generation module} that tackles task (i) and a \textit{hierarchy proposal module} that handles tasks (ii) and (iii) (see Figure~\ref{fig:pipeline})). Additionally, our pipeline can generate multiple (up to five) potential hierarchies per topic. We describe our pipeline module in further detail below and provide complete prompt details in Appendix \ref{sec:prompt}.

\subsubsection{Pre-Generation Module}
This module extracts relevant content from a set of studies to use as input for hierarchy proposal. 

\para{Claim generation.} We generate concise claim statements from a given scientific study to reduce the amount of information provided as input to subsequent LLM modules. Providing a study abstract as input, we prompt a LLM to generate claims describing all findings discussed. We qualitatively examine the claim generation capabilities of two state-of-the-art LLMs: (i) \textsc{GPT-3.5} (June 2023 version) and (ii) \textsc{Claude-2}. Our assessment indicates that GPT-3.5 performs better in terms of clarity and conciseness; therefore we extract claim statements for all studies in our dataset using this model. Additionally, to assess whether generated claims contain hallucinated information, we run a fine-tuned \textsc{Deberta-V3} NLI model \cite{Laurer_van2024} on abstract-claim pairs. 
We observe that 98.1\% of the generated claims are entailed by their corresponding study abstracts, indicating that claims are generally faithful to source abstracts.\footnote{We further conduct a qualitative evaluation to ensure factuality of generated claims in Appendix Section \ref{sec:claim-fact}.} 
These sets of generated claims are provided as input to the hierarchy proposal module.
\paragraph{Frequent entity extraction.} 
Based on preliminary exploration, we observe that simply prompting LLMs to generate hierarchies given a set of claims often produces hierarchy categories with low coverage over the claim set. Therefore, we extract frequently-occurring entities to provide as additional cues to bias category generation. We use \textsc{ScispaCy} \cite{neumann-etal-2019-scispacy} to extract entities from all study abstracts, then aggregate and sort them by frequency. The 20 most frequent entities are used as additional keywords to bias generated categories towards having high coverage. 

\subsubsection{Hierarchy Proposal Module}
The aim of this module is to generate final hierarchies in two steps within a single prompt: (i) generate possible categories that can form the root node of a hierarchy (i.e., categories that divide claims into various clusters), and then (ii) generate the complete hierarchy with claim organization. For instance, considering the example in Figure~\ref{fig:example}, step (i) would produce root categories ``exercise modalities'' and ``cancer types'' and step (ii) would produce all sub-categories ($S1-S5$) and organize studies under them (e.g., assigning studies 1, 3, 5 under S1).

\para{Root category generation.}
With outputs from the pre-generation module and a research topic (systematic review title), we prompt the LLM to generate up to five top-level aspects as possible root categories for hierarchies.

\para{Hierarchy completion.}
With generated root categories, this step aims to produce a complete hierarchy. We prompt the LLM to produce one hierarchy per root category, with every non-root category also containing numeric references to claims categorized under it. Note that in our setting, a claim may be assigned to multiple categories or remain uncategorized. 
A manual comparison of \textsc{GPT-4} and \textsc{Claude-2} outputs shows that \textsc{Claude-2} generates deeper hierarchies compared to \textsc{GPT-4}, which only generate shallow hierarchies
, so we use \textsc{Claude-2} for the hierarchy proposal module. 
See more details in Appendix Section \ref{sec:model-selection}.

Using this pipeline, we generate 2,174 preliminary hierarchies ($\sim$4.6 hierarchies per review) for our curated set of 472 systematic reviews (or sets of related studies).

\section{Correcting Hierarchies via Human Feedback}
The second phase of our dataset creation process involves correction of preliminary LLM-generated hierarchies via human feedback. Correcting these hierarchies is challenging because of two issues. First, the volume of information present in generated hierarchies (links between categories, claim-category links, etc.) makes correction very time-consuming, especially in a single pass. Second, since categories and claims in a hierarchy are inter-linked, corrections can have cascading effects (e.g., changing a category name can affect which claims should be categorized under it). These issues motivate us to decompose hierarchy correction into three sub-tasks, making the feedback process less tedious and time-consuming. Furthermore, each sub-task focuses on the correction of only one category of links to mitigate cascading effects. These three sub-tasks are: (i) assessing correctness of parent-child category links, (ii) assessing coherence of sibling category groups, and (iii) assessing claim categorization.

\subsection{Assessing Parent-Child Category Links}
In this sub-task, given all parent-child category links from a hierarchy (e.g., $P1\rightarrow S[1-5]$ in Figure~\ref{fig:example}), for each link, humans are prompted to determine whether the child is a valid sub-category of the parent. Annotators can label parent-child category links using one of the following labels: (i) parent and child categories have a hypernym-hyponym relationship (e.g., exercise modalities $\rightarrow$ aerobic exercise), (ii) parent and child categories are not related by hypernymy but the child category provides a useful breakdown of the parent(e.g., aerobic exercise $\rightarrow$ positive effects), and (iii) parent and child categories are unrelated (e.g., aerobic exercise $\rightarrow$ anaerobic exercise). Categories (i) and (ii) are positive labels indicating valid links, while category (iii) is a negative label capturing incorrect links in the existing hierarchy.

\subsection{Assessing Coherence of Sibling Categories}
For a hierarchical organization to be useful, in addition to validity of parent-child category links, all sibling categories (i.e., categories under the same parent, like $S1-S5$ in Figure~\ref{fig:example}) should also be coherent. Therefore, in our second sub-task, given a parent and all its child categories, we ask annotators to determine whether these child categories form a coherent sibling group. Annotators can assign a positive or negative coherence label to each child category in the group. For example, given the parent category ``type of cancer'' and the set of child categories ``liver cancer'', ``prostate cancer'', ``lung cancer'', and ``recurrence'', the first three categories are assigned positive labels, while ``recurrence'' is assigned a negative label since it is not a type of cancer. All categories assigned a negative label capture incorrect groups in the existing hierarchy.

\subsection{Assessing Claim Categorization}
Unlike the previous sub-tasks which focus on assessing links between categories at all levels of the hierarchy, the final sub-task focuses on assessing the assignment of claims to various categories. Given a claim and \emph{all} categories present in the hierarchy, for each claim-category pair, humans are prompted to assess whether the claim contains any information relevant to that category. The claim-category pair is assigned a positive label if relevant information is present, and negative otherwise. For every category, we include the path from the root to provide additional context which might be needed to interpret it accurately (e.g., ``positive findings'' has a broader interpretation than ``chemotherapy $\rightarrow$ positive findings''). Instead of only assessing relevance of categories under which a claim has currently been categorized, this sub-task evaluates all claim-category pairs in order to catch recall errors, i.e., cases in which a claim could be assigned to an category but is not categorized there currently.

\subsection{Feedback Process}
\paragraph{Data Sampling:} 
Due to the time-intensiveness of the correction task, we collect annotations for 100 / 472 randomly-sampled topics, and further filter out hierarchies which cover less than 30\% of the claims associated with that topic. This leaves us with 320 hierarchies to collect corrections for. For the parent-child link assessment sub-task, this produces 1,635 links to be assessed. For sibling coherence, after removing all parent categories with only one child, we obtain 574 sibling groups to be assessed. Lastly, for claim categorization, the most intensive task, we end up with 50,723 claim-category pairs to label.

\paragraph{Annotator Background:} We recruit a team of five experts with backgrounds in biology or medicine to conduct annotations. Two of these experts are authors on this paper, and the remaining three were recruited via Upwork.\footnote{\url{https://www.upwork.com/}} 
Every annotator is required to first complete a qualification test, which includes sample data from all three sub-tasks, and must achieve reasonable performance before they are asked to annotate data. 

\paragraph{Annotation Pilots:} Given the complexity and ambiguity of our tasks, we conduct several rounds of pilot annotation with iterative feedback before commencing full-scale annotation. This ensures that all annotators develop a deep understanding of the task and can achieve high agreement.
After each pilot, we measure inter-annotator agreement on each sub-task. Due to the presence of unbalanced labels in tasks 1 and 2, we compute agreement using match rate; for task 3, we report Fleiss' kappa. At the end of all pilot rounds, we achieve high agreement on all sub-tasks, with match rates of 100\% and 78\% on tasks 1 and 2 respectively and Fleiss' kappa of 0.66 on task 3.

\begin{table}[]
\centering
\begin{tabular}{@{}lccc@{}}
\toprule
              & Precision      & Recall         & F1            \\ \midrule
Task 1      & 0.999       & -   & -        \\
Task 2      &   0.773       & -    &-          \\
Task 3      & 0.716          & 0.539    & 0.615         \\\bottomrule
\end{tabular}
\caption{
Performance assessment of our LLM-based pipeline on expert-curated hierarchies for 100 topics. Recall and F1 for tasks 1 and 2 cannot be measured since we only get positive predictions from the pipeline.
}
\label{tab:pipeline-result}
\end{table}

\subsection{Assessment of Preliminary Hierarchies}
\label{sec:hier_assess}
An additional benefit of collecting corrections for preliminary hierarchies (as described above) is that this data allows us to quantify the quality of our LLM-generated hierarchies and measure the performance of our hierarchy generation pipeline.

\paragraph{Parent-child link accuracy.} Interestingly, we observe almost perfect performance on this sub-task, with only one out of 1635 parent-child links being labeled as incorrect where the pipeline put ``Coffee consumption'' under ``Tea consumption and cancer risk''. Of the remaining correct links, 75\% are labeled as hypernym-hyponym links, and 25\% as useful breakdowns of the parent category. This result demonstrates that LLMs are highly accurate at generating good sub-categories given a parent category, even when dealing with long inputs.

\paragraph{Sibling coherence performance.} Next, we look into LLM performance on sibling coherence and observe that this is also fairly high, with 77\% of sibling groups being labeled as coherent where ``coherent'' denotes a sibling group in which expert labels for all sibling categories are positive; otherwise, ``zero.'' Among sibling groups labeled incoherent, we observe two common types of errors: (1) categories at different levels of granularity being grouped as siblings, and (2) one or more categories having subtly different focuses. For example, \figref{fig:example} demonstrates a type 1 error, where the sub-category ``walking'' is more specific and should be classified under ``aerobic'' but is instead listed as a sibling. An example of a type 2 error is the parent category ``dietary interventions'' with child categories ``low calorie diets'', ``high/low carbohydrate diets'', and ``prepared meal plans''. Here, though all child categories are dietary interventions, the first two have an explicit additional focus on nutritional value which ``prepared meal plans'' lacks, making them incoherent as a sibling group. 

\paragraph{Claim categorization performance.} The design of our claim categorization sub-task prompts annotators to evaluate the relationship between a given claim and every category in the hierarchy. Hence, when assessing whether annotators agree with the LLM's categorization of a claim under a category, we need to aggregate over the labels assigned to all claim-category pairs from the root to the target category under consideration. Formally, for a claim-category pair $(cl_i, ct_j)$, instead of only using label $l_{ij} = h((cl_i, ct_j))$ from human feedback $h$, we must aggregate over labels assigned to all ancestors of $ct_j$, i.e., $L = [h((cl_i, ct_1)), ..., h((cl_i, ct_j))]$, where $ct_1$ is the root category and $ct_j$ is the target category. We do this aggregation using an AND operation $l_{agg} = l_1 \land l_2 \land ... \land l_j$. After computing these aggregate labels, we observe that our LLM-based pipeline has reasonable precision (0.71), but much lower recall (0.53) on claim categorization. A low recall rate on this sub-task is problematic because, while it is easy for human annotators to correct precision errors (remove claims wrongly assigned to various categories), it is much harder to correct recall errors (identify which claims were missed under a given category), which necessitates a thorough examination of all studies.

\section{Characterizing Hierarchy Complexity}
\label{sec:hierarchychar}
Our dataset creation process produces 2,174 hierarchies on 472 research topics, with 320 hierarchies (for 100 topics) corrected by domain experts. We briefly characterize the complexity of all generated hierarchies, focusing on two aspects: (i) structural complexity, and (ii) semantic complexity.

\subsection{Structural Complexity}
\para{Hierarchy depth:} All generated hierarchies are multi-level, with a mean hierarchy depth of 2.5, and maximum depth of 5.

\para{Node arity:} On average, every parent has a node arity of 2.4 (i.e., has 2.4 child categories). However, node arity can grow as large as 10 for certain parent categories.

\para{Claim coverage:} Another crucial property of generated hierarchies is their coverage of claims since hierarchies containing fewer claims are easier to generate but less useful. We observe that given a set of claims, a typical hierarchy incorporates 12.3 claims on average. Additionally, very few claims from a set remain uncategorized, i.e., not covered by any generated hierarchy (2.6 on average). 

These characteristics indicate that our LLM-generated hierarchies have interesting structural properties.

\subsection{Semantic Complexity}
\para{Category diversity:} Our dataset contains 4.6 hierarchies per research topic. We manually inspect a small sample of hierarchies for 10 research topics, and find that none of the hierarchies generated for a single topic contain any repeating categories. This signals that the multiple hierarchies we generate per topic represent semantically diverse ways of grouping/slicing the same set of claims.

\para{Adherence to PICO framework:} Systematic reviews in biomedicine typically use the PICO (population, intervention, comparator, outcome) framework \cite{richardson1995well} to categorize studies. To understand how much our generated hierarchies adhere to this framework, we again inspect hierarchies for 10 research topics and label whether the root category focuses on a PICO element. We observe that 34 out of 46 hierarchies have a PICO-focused root category, making them directly useful for systematic review. Interestingly, the remaining hierarchies still focus on useful categories such as continuing patient education, study limitations, cost analyses etc. Thus, besides surfacing categorizations expected by the systematic review process, using LLMs can help discover additional interesting categorizations.

\section{Automating Hierarchy Correction}
As mentioned in \S\ref{sec:hierarchychar}, we hire five domain experts to correct hierarchies for 100 research topics.
However, the correction process, despite our best efforts at task simplification and decomposition, is still time-consuming and requires domain expertise. Therefore, we investigate whether we can use our corrected hierarchy data to automate some correction sub-tasks. In particular, we focus on automating sibling coherence and claim categorization correction since Table~\ref{tab:pipeline-result} indicates that LLMs already achieve near-perfect performance on producing relevant child categories for a parent. 

\begin{table}[]
\centering
\begin{tabular}{@{}lrrr@{}}
\toprule
           & Task 1 & Task 2 & Task 3 \\ \midrule
Train      & 838    & 298    & 23,692 \\
Validation & 285    & 99     & 8,241  \\
Test ID    & 327    & 115    & 13,595 \\
Test OOD   & 185    & 62     & 5,195  \\ \midrule
Total      & 1,635  & 574    & 50,723 \\ \bottomrule
\end{tabular}
\caption{Dataset statistics for three correction sub-tasks: 
parent-child category links (Task 1), sibling category coherence (Task 2), and claim categorization (Task 3).
}
\label{tab:dataset-stats}
\end{table}

\subsection{Experimental Setup}
We briefly discuss the experimental setup we use to evaluate whether model performance on sibling coherence and claim categorization correction can be improved using our collected feedback data.

\subsubsection{Dataset Split}
To better assess generalizability, we carefully construct two test sets, an in-domain (ID) and an out-of-domain (OOD) subset instead of randomly splitting our final dataset of 100 research topics. To develop our OOD test set, we first embed all 100 research topics by running \texttt{SPECTER2}, 
a scientific paper embedding model using citation graph
\cite{Singh2022SciRepEvalAM}, on the title and abstract of the Cochrane systematic review associated with each topic. Then, we run hierarchical clustering on the embeddings and choose one isolated cluster ($n=12$ reviews) to be our OOD test set. Our manual inspection reveals that all studies in this cluster are about \textit{fertility} and \textit{pregnancy}. After creating our OOD test set, we then randomly sub-sample our ID test set ($n=18$ reviews) from remaining research topics. This leaves us with 70 topics, which we split into training and validation sets. Detailed statistics for our dataset splits, including number of instances for each correction sub-task, are provided in \tableref{tab:dataset-stats}.

\begin{table*}[]
\centering
\small
\begin{tabular}{@{}llccccccccc@{}}
\toprule
& & \multicolumn{3}{c}{All}                          & \multicolumn{3}{c}{In-domain}                    & \multicolumn{3}{c}{Out-of-domain}                \\ 
& & Precision      & Recall         & F1             & Precision      & Recall         & F1             & Precision      & Recall         & F1             \\ \midrule
\multirow{2}{*}{\textbf{Fine-tuned}} & Flan-T5 base      & 0.368          & 0.179          & 0.241          & 0.364          & 0.167          & 0.229          & 0.375          & 0.200          & 0.261          \\
    & Flan-T5 large     & 0.333          & 0.333          & 0.333          & 0.269          & 0.292          & 0.280           & \textbf{0.462} & 0.400          & 0.429          \\\midrule
\multirow{2}{*}{\shortstack[l]{\textbf{Zero-shot}\\\textbf{CoT}}} & GPT-3.5 Turbo & 0.419          & \textbf{0.667} & \textbf{0.515} & 0.400          & \textbf{0.583} & \textbf{0.475} & 0.444          & \textbf{0.800} & \textbf{0.571} \\
& GPT-4 Turbo         & \textbf{0.467} & 0.359          & 0.406          & \textbf{0.474} & 0.375          & 0.419          & 0.455          & 0.333          & 0.385          \\ \bottomrule
\end{tabular}
\caption{Performance of all models on assessing sibling coherence. }
\label{tab:task2-correct}
\end{table*}

\begin{table*}[]
\centering
\small
\begin{tabular}{@{}llccccccccc@{}}
\toprule
      & & \multicolumn{3}{c}{All}                          & \multicolumn{3}{c}{In-domain}                    & \multicolumn{3}{c}{Out-of-domain}                \\ 
              & & Precision      & Recall         & F1             & Precision      & Recall         & F1             & Precision      & Recall         & F1             \\ \midrule
& Pipeline      & 0.697          & {\ul 0.567}    & 0.625          & 0.677          & {\ul 0.575}    & 0.622          & 0.757          & {\ul 0.548}    & 0.636          \\\midrule
\multirow{2}{*}{\textbf{Fine-tuned}} & Flan-T5-base  & 0.767          & 0.711          & 0.738          & 0.750           & 0.702          & 0.725          & \textbf{0.816} & 0.735 & \textbf{0.773} \\
& Flan-T5-large & \textbf{0.779} & 0.726 & \textbf{0.751} & \textbf{0.769} & 0.726 & \textbf{0.747} & 0.807          & 0.725          & 0.764          \\\midrule
\multirow{2}{*}{\shortstack[l]{\textbf{Zero-shot}\\\textbf{CoT}}} & GPT-3.5 Turbo & 0.585          & 0.861 & 0.697 & 0.570          & 0.871 & 0.689 & 0.631          & 0.835 & 0.719 \\
& GPT-4 Turbo        & 0.557     & \textbf{0.932} & 0.697          & 0.544     & \textbf{0.933} & 0.687          & 0.594          & \textbf{0.932} & 0.726          \\ \bottomrule
\end{tabular}
\caption{
Performance of all models on correcting claim categorization.
}
\label{tab:task3-full}
\end{table*}

\subsubsection{Models}
We evaluate two classes of methods for correction:
\begin{itemize}[leftmargin=*,topsep=0pt]
\setlength\itemsep{-0.5em}
    \item \textbf{Finetuned LMs:} To assess whether correction abilities of smaller LMs can be improved by finetuning on our collected feedback data, we experiment with Flan-T5  \cite{Chung2022ScalingIL}, which has proven to be effective on many benchmarks. 
    \item \textbf{Zero-Shot CoT:} To explore whether using chain-of-thought (CoT) prompting \cite{Wei2022ChainOT} improves the ability of LLMs to do correction zero-shot without using our feedback data, we test OpenAI GPT-3.5 Turbo (\texttt{gpt-3.5-turbo-0613}) and GPT-4 Turbo (\texttt{gpt-4-1106-preview}).
\end{itemize}

\noindent Additional modeling details including CoT prompts are provided in Appendix \ref{appx:model}.

\subsection{Correcting Sibling Coherence}
Table~\ref{tab:task2-correct} presents the performance of all models on the task of identifying sibling groups that are incoherent. Finetuning models on this task is challenging due to the small size of the training set ($n=298$) and imbalanced labels. Despite upsampling and model selection based on precision, finetuned Flan-T5 models do not perform well on this task (best F1-score of $33.3\%$). Additionally LLMs also do not perform well despite the use of chain-of-thought prompting to handle the complex reasoning required for this task. At $51.5\%$ F1, LLMs outperform finetuned models; however, their precision ($46.7\%$ for \texttt{GPT-4-Turbo}) is still not good enough to detect incoherent sibling groups confidently. 
These results indicate that this correction sub-task is extremely difficult to automate and will likely continue to require expert intervention.

\subsection{Correcting Claim Categorization}
Table~\ref{tab:task3-full} shows the performance of all models on the task of correcting assignment of claims to categories in the hierarchy. Following the strategy described in \S\ref{sec:hier_assess}, given a claim, we first use our models to generate predictions for every claim-category pair (all category nodes) and then obtain the final label for each category by applying an AND operation over all predictions from the root category to that category. Our results show that this task is easier to automate---fine-tuning Flan-T5 on our collected training dataset leads to better scores on all metrics compared to our LLM pipeline. Crucially, recall which is much more time-consuming for humans to fix, improves by $15.9$ points using Flan-T5-large indicating that automating this step can provide additional efficiency gains during correction. LLMs perform well too, with GPT-4-Turbo achieving the best recall rate among all models, but its lower precision score makes the predictions less reliable overall.

Interestingly, we notice that all models perform better on the OOD test for both correction tasks, indicating that the OOD test set likely contains instances that are less challenging than the ID set.

\subsection{Correcting Claim Categorization for Remaining Hierarchies}
Comparing the claim categorization predictions of Flan-T5-large on our test set with our LLM-based hierarchy generation pipeline reveals that it flips labels in $24.7\%$ cases, of which 63.5\% changes are correct. This indicates that a FLAN-T5-large corrector can potentially improve claim categorization of LLM-generated hierarchies. Therefore, we apply this corrector to the remaining 372 LLM-generated hierarchies that we do not have expert corrections for to improve claim assignment for those. Our final curated dataset \dataset{} contains hierarchies for 472 research topics, of which hierarchies for 100 topics have been corrected by experts on both category linking and claim categorization, while hierarchies for the remaining 372 have had claim assignments corrected automatically.

\section{Related Work}
\subsection{Literature Review Support}
Prior work on developing literature review support tools has largely focused on using summarization techniques for end-to-end review generation or to tackle specific aspects of the problem (see \citet{altmami2022automatic} for a detailed survey). Some studies have focused on generating ``citation sentences'' discussing relationships between related papers, which can be included in a literature review \cite{xing-etal-2020-automatic,luu-etal-2021-explaining,ge-etal-2021-baco,wu2021towards}. Other work has focused on the task of generating related work sections for a scientific paper \cite{hoang-kan-2010-towards,hu-wan-2014-automatic,li2022generating,wang2022multi}, which while similar in nature to literature review, has a narrower scope and expects more concise generation outputs. Finally, motivated by the ever-improving capabilities of generative models, some prior work has attempted to automate end-to-end review generation treating it as multi-document summarization, with limited success \cite{mohammad-etal-2009-using,jha-etal-2015-content,Wallace2020GeneratingN,deyoung-etal-2021-ms,liu2023generating,zhu-etal-2023-hierarchical}. Of these, \citet{zhu-etal-2023-hierarchical} generates intermediate \textit{hierarchical outlines} to scaffold literature review generation, but unlike our work, they do not produce multiple organizations for the same set of related studies. Additionally, we focus solely on the problem of organizing related studies for literature review, leaving review generation and writing assistance to future work. 

\subsection{LLMs for Organization} 
Organizing document collections is an extensively-studied problem in NLP, with several classes of approaches such as clustering and topic modeling \cite{dumais1988using} addressing this goal. Despite their utility, conventional clustering and topic modeling approaches are not easily interpretable \cite{chang2009reading}, requiring manual effort which introduces subjectivity and affects their reliability \cite{baden2022three}. Recent work has started exploring whether using LLMs for clustering \cite{viswanathan2023large, Zhang2023ClusterLLMLL,Wang2023GoalDrivenEC} and topic modeling \cite{Pham2023TopicGPTAP} can alleviate some of these issues, with promising results. This motivates us to experiment with LLMs for generating hierarchical organizations of scientific studies. Interestingly, TopicGPT \cite{Pham2023TopicGPTAP} also attempts to perform hierarchical topic modeling, but is limited to producing two-level hierarchies unlike our approach which generates hierarchies of arbitrary depth.

\section{Conclusion}
Our work explored the utility of LLMs for producing hierarchical organizations of scientific studies, with the goal of assisting researchers in performing literature review. We collected \dataset{}, an expert-curated dataset for hierarchy generation focused on biomedicine, using a human-in-the-loop process in which a naive LLM-based pipeline generates preliminary hierarchies which are corrected by experts. To make hierarchy correction less tedious and time-consuming, we decomposed it into a three-step process in which experts assessed the correctness of links between categories as well as assignment of studies to categories. \dataset{} contains 2,174 LLM-generated hierarchies covering 472 topics, and expert-corrected hierarchies for a subset of 100 topics. Quantifying LLM performance using our collected data revealed that LLMs are quite good at generating and linking categories, but needed further improvement on study assignment. We trained a corrector model with our feedback data which improved study assignment further by $12.6\%$ F1 points. We hope that releasing \dataset{} and our hierarchy generation and correction models will motivate further research on developing better assistive tools for literature review.

\section*{Limitations}
\paragraph{Single-domain focus.} Given our primary focus on biomedicine, it is possible that our hierarchy generation and correction methods do not generalize well to other scientific domains. Further investigation of generalization is out of scope for this work but a promising area for future research.

\paragraph{Deployment difficulties.} Powerful LLMs like \textsc{Claude-2} have long inference times --- in some cases, the entire hierarchy generation process can take up to one minute to complete. This makes it extremely challenging to deploy our hierarchy construction pipeline as a real-time application. However, it is possible to conduct controlled lab studies to evaluate the utility of our pipeline as a literature review assistant, which opens up another line of investigation for future work.  

\paragraph{Reliance on curated sets of related studies.} Our current hierarchical organization pipeline relies on the assumption that all provided studies are relevant to the research topic being reviewed. 
However, in a realistic literature review setting, researchers often retrieve a set of studies from search engines, which may or may not be relevant to the topic of interest, and are interested in organizing the retrieved results. 
In a preliminary qualitative analysis in Appendix Section \ref{sec:retrieval-quality}, we show that our system can handle some noise in retrieved studies, though we defer a detailed robustness evaluation to future work.

\section*{Acknowledgements}
We would like to thank the reviewers, Joseph Chee Chang, and the rest of the Semantic Scholar team at AI2 for their valuable feedback and comments. We also want to thank the Upworkers who participated in our formative studies and annotation process.

\bibliography{anthology,custom}

\begin{thebibliography}{36}
\expandafter\ifx\csname natexlab\endcsname\relax\def\natexlab#1{#1}\fi

\bibitem[{Altmami and Menai(2022)}]{altmami2022automatic}
Nouf~Ibrahim Altmami and Mohamed El~Bachir Menai. 2022.
\newblock Automatic summarization of scientific articles: A survey.
\newblock \emph{Journal of King Saud University-Computer and Information
  Sciences}, 34(4):1011--1028.

\bibitem[{Baden et~al.(2022)Baden, Pipal, Schoonvelde, and van~der
  Velden}]{baden2022three}
Christian Baden, Christian Pipal, Martijn Schoonvelde, and Mariken AC~G van~der
  Velden. 2022.
\newblock Three gaps in computational text analysis methods for social
  sciences: A research agenda.
\newblock \emph{Communication Methods and Measures}, 16(1):1--18.

\bibitem[{Borah et~al.(2017)Borah, Brown, Capers, and
  Kaiser}]{borah2017analysis}
Rohit Borah, Andrew~W Brown, Patrice~L Capers, and Kathryn~A Kaiser. 2017.
\newblock Analysis of the time and workers needed to conduct systematic reviews
  of medical interventions using data from the prospero registry.
\newblock \emph{BMJ open}, 7(2).

\bibitem[{Chang et~al.(2009)Chang, Gerrish, Wang, Boyd-Graber, and
  Blei}]{chang2009reading}
Jonathan Chang, Sean Gerrish, Chong Wang, Jordan Boyd-Graber, and David Blei.
  2009.
\newblock Reading tea leaves: How humans interpret topic models.
\newblock \emph{Advances in neural information processing systems}, 22.

\bibitem[{Chung et~al.(2022)Chung, Hou, Longpre, Zoph, Tay, Fedus, Li, Wang,
  Dehghani, Brahma, Webson, Gu, Dai, Suzgun, Chen, Chowdhery, Valter, Narang,
  Mishra, Yu, Zhao, Huang, Dai, Yu, Petrov, hsin Chi, Dean, Devlin, Roberts,
  Zhou, Le, and Wei}]{Chung2022ScalingIL}
Hyung~Won Chung, Le~Hou, S.~Longpre, Barret Zoph, Yi~Tay, William Fedus, Eric
  Li, Xuezhi Wang, Mostafa Dehghani, Siddhartha Brahma, Albert Webson,
  Shixiang~Shane Gu, Zhuyun Dai, Mirac Suzgun, Xinyun Chen, Aakanksha
  Chowdhery, Dasha Valter, Sharan Narang, Gaurav Mishra, Adams~Wei Yu, Vincent
  Zhao, Yanping Huang, Andrew~M. Dai, Hongkun Yu, Slav Petrov, Ed~Huai hsin
  Chi, Jeff Dean, Jacob Devlin, Adam Roberts, Denny Zhou, Quoc~V. Le, and Jason
  Wei. 2022.
\newblock \href {https://api.semanticscholar.org/CorpusID:253018554} {Scaling
  instruction-finetuned language models}.
\newblock \emph{ArXiv}, abs/2210.11416.

\bibitem[{D'Arcy et~al.(2024)D'Arcy, Hope, Birnbaum, and Downey}]{d2024marg}
Mike D'Arcy, Tom Hope, Larry Birnbaum, and Doug Downey. 2024.
\newblock Marg: Multi-agent review generation for scientific papers.
\newblock \emph{arXiv preprint arXiv:2401.04259}.

\bibitem[{DeYoung et~al.(2021)DeYoung, Beltagy, van Zuylen, Kuehl, and
  Wang}]{deyoung-etal-2021-ms}
Jay DeYoung, Iz~Beltagy, Madeleine van Zuylen, Bailey Kuehl, and Lucy~Lu Wang.
  2021.
\newblock \href {https://doi.org/10.18653/v1/2021.emnlp-main.594} {{MS}{\^{}}2:
  Multi-document summarization of medical studies}.
\newblock In \emph{Proceedings of the 2021 Conference on Empirical Methods in
  Natural Language Processing}, pages 7494--7513, Online and Punta Cana,
  Dominican Republic. Association for Computational Linguistics.

\bibitem[{Dumais et~al.(1988)Dumais, Furnas, Landauer, Deerwester, and
  Harshman}]{dumais1988using}
Susan~T Dumais, George~W Furnas, Thomas~K Landauer, Scott Deerwester, and
  Richard Harshman. 1988.
\newblock Using latent semantic analysis to improve access to textual
  information.
\newblock In \emph{Proceedings of the SIGCHI conference on Human factors in
  computing systems}, pages 281--285.

\bibitem[{Ge et~al.(2021)Ge, Dinh, Liu, Su, Lu, Wang, and
  Diesner}]{ge-etal-2021-baco}
Yubin Ge, Ly~Dinh, Xiaofeng Liu, Jinsong Su, Ziyao Lu, Ante Wang, and Jana
  Diesner. 2021.
\newblock \href {https://doi.org/10.18653/v1/2021.acl-long.116} {{BACO}: A
  background knowledge- and content-based framework for citing sentence
  generation}.
\newblock In \emph{Proceedings of the 59th Annual Meeting of the Association
  for Computational Linguistics and the 11th International Joint Conference on
  Natural Language Processing (Volume 1: Long Papers)}, pages 1466--1478,
  Online. Association for Computational Linguistics.

\bibitem[{Hoang and Kan(2010)}]{hoang-kan-2010-towards}
Cong Duy~Vu Hoang and Min-Yen Kan. 2010.
\newblock \href {https://aclanthology.org/C10-2049} {Towards automated related
  work summarization}.
\newblock In \emph{Coling 2010: Posters}, pages 427--435, Beijing, China.
  Coling 2010 Organizing Committee.

\bibitem[{Hu and Wan(2014)}]{hu-wan-2014-automatic}
Yue Hu and Xiaojun Wan. 2014.
\newblock \href {https://doi.org/10.3115/v1/D14-1170} {Automatic generation of
  related work sections in scientific papers: An optimization approach}.
\newblock In \emph{Proceedings of the 2014 Conference on Empirical Methods in
  Natural Language Processing ({EMNLP})}, pages 1624--1633, Doha, Qatar.
  Association for Computational Linguistics.

\bibitem[{Jha et~al.(2015)Jha, Finegan-Dollak, King, Coke, and
  Radev}]{jha-etal-2015-content}
Rahul Jha, Catherine Finegan-Dollak, Ben King, Reed Coke, and Dragomir Radev.
  2015.
\newblock \href {https://doi.org/10.3115/v1/P15-1043} {Content models for
  survey generation: A factoid-based evaluation}.
\newblock In \emph{Proceedings of the 53rd Annual Meeting of the Association
  for Computational Linguistics and the 7th International Joint Conference on
  Natural Language Processing (Volume 1: Long Papers)}, pages 441--450,
  Beijing, China. Association for Computational Linguistics.

\bibitem[{Kojima et~al.(2022)Kojima, Gu, Reid, Matsuo, and
  Iwasawa}]{kojima2022large}
Takeshi Kojima, Shixiang~Shane Gu, Machel Reid, Yutaka Matsuo, and Yusuke
  Iwasawa. 2022.
\newblock Large language models are zero-shot reasoners.
\newblock \emph{Advances in neural information processing systems},
  35:22199--22213.

\bibitem[{Laurer et~al.(2024)Laurer, van Atteveldt, Casas, and
  Welbers}]{Laurer_van2024}
Moritz Laurer, Wouter van Atteveldt, Andreu Casas, and Kasper Welbers. 2024.
\newblock \href {https://doi.org/10.1017/pan.2023.20} {Less annotating, more
  classifying: Addressing the data scarcity issue of supervised machine
  learning with deep transfer learning and bert-nli}.
\newblock \emph{Political Analysis}, 32(1):84–100.

\bibitem[{Li et~al.(2022)Li, Lu, and Cheng}]{li2022generating}
Pengcheng Li, Wei Lu, and Qikai Cheng. 2022.
\newblock Generating a related work section for scientific papers: an optimized
  approach with adopting problem and method information.
\newblock \emph{Scientometrics}, 127(8):4397--4417.

\bibitem[{Liu et~al.(2023)Liu, Lin, Hewitt, Paranjape, Bevilacqua, Petroni, and
  Liang}]{liu2023lost}
Nelson~F. Liu, Kevin Lin, John Hewitt, Ashwin Paranjape, Michele Bevilacqua,
  Fabio Petroni, and Percy Liang. 2023.
\newblock \href {https://api.semanticscholar.org/CorpusID:259360665} {Lost in
  the middle: How language models use long contexts}.
\newblock \emph{Transactions of the Association for Computational Linguistics},
  12:157--173.

\bibitem[{Liu et~al.(2022)Liu, Cao, Yang, and Wen}]{liu2023generating}
Shuaiqi Liu, Jiannong Cao, Ruosong Yang, and Zhiyuan Wen. 2022.
\newblock \href {https://api.semanticscholar.org/CorpusID:250636132}
  {Generating a structured summary of numerous academic papers: Dataset and
  method}.
\newblock In \emph{International Joint Conference on Artificial Intelligence}.

\bibitem[{Luu et~al.(2021)Luu, Wu, Koncel-Kedziorski, Lo, Cachola, and
  Smith}]{luu-etal-2021-explaining}
Kelvin Luu, Xinyi Wu, Rik Koncel-Kedziorski, Kyle Lo, Isabel Cachola, and
  Noah~A. Smith. 2021.
\newblock \href {https://doi.org/10.18653/v1/2021.acl-long.166} {Explaining
  relationships between scientific documents}.
\newblock In \emph{Proceedings of the 59th Annual Meeting of the Association
  for Computational Linguistics and the 11th International Joint Conference on
  Natural Language Processing (Volume 1: Long Papers)}, pages 2130--2144,
  Online. Association for Computational Linguistics.

\bibitem[{Mohammad et~al.(2009)Mohammad, Dorr, Egan, Hassan, Muthukrishan,
  Qazvinian, Radev, and Zajic}]{mohammad-etal-2009-using}
Saif Mohammad, Bonnie Dorr, Melissa Egan, Ahmed Hassan, Pradeep Muthukrishan,
  Vahed Qazvinian, Dragomir Radev, and David Zajic. 2009.
\newblock \href {https://aclanthology.org/N09-1066} {Using citations to
  generate surveys of scientific paradigms}.
\newblock In \emph{Proceedings of Human Language Technologies: The 2009 Annual
  Conference of the North {A}merican Chapter of the Association for
  Computational Linguistics}, pages 584--592, Boulder, Colorado. Association
  for Computational Linguistics.

\bibitem[{Neumann et~al.(2019)Neumann, King, Beltagy, and
  Ammar}]{neumann-etal-2019-scispacy}
Mark Neumann, Daniel King, Iz~Beltagy, and Waleed Ammar. 2019.
\newblock \href {https://doi.org/10.18653/v1/W19-5034} {{S}cispa{C}y: Fast and
  robust models for biomedical natural language processing}.
\newblock In \emph{Proceedings of the 18th BioNLP Workshop and Shared Task},
  pages 319--327, Florence, Italy. Association for Computational Linguistics.

\bibitem[{Pham et~al.(2023)Pham, Hoyle, Sun, and Iyyer}]{Pham2023TopicGPTAP}
Chau~Minh Pham, Alexander~Miserlis Hoyle, Simeng Sun, and Mohit Iyyer. 2023.
\newblock \href {https://api.semanticscholar.org/CorpusID:264935207} {Topicgpt:
  A prompt-based topic modeling framework}.
\newblock \emph{ArXiv}, abs/2311.01449.

\bibitem[{Richardson et~al.(1995)Richardson, Wilson, Nishikawa, and
  Hayward}]{richardson1995well}
W~Scott Richardson, Mark~C Wilson, Jim Nishikawa, and Robert~S Hayward. 1995.
\newblock The well-built clinical question: a key to evidence-based decisions.
\newblock \emph{ACP journal club}, 123(3):A12--A13.

\bibitem[{Shojania et~al.(2007)Shojania, Sampson, Ansari, Ji, Doucette, and
  Moher}]{shojania2007quickly}
Kaveh~G Shojania, Margaret Sampson, Mohammed~T Ansari, Jun Ji, Steve Doucette,
  and David Moher. 2007.
\newblock How quickly do systematic reviews go out of date? a survival
  analysis.
\newblock \emph{Annals of internal medicine}, 147(4):224--233.

\bibitem[{Singh et~al.(2022)Singh, D'Arcy, Cohan, Downey, and
  Feldman}]{Singh2022SciRepEvalAM}
Amanpreet Singh, Mike D'Arcy, Arman Cohan, Doug Downey, and Sergey Feldman.
  2022.
\newblock \href {https://api.semanticscholar.org/CorpusID:254018137}
  {Scirepeval: A multi-format benchmark for scientific document
  representations}.
\newblock In \emph{Conference on Empirical Methods in Natural Language
  Processing}.

\bibitem[{Viswanathan et~al.(2023)Viswanathan, Gashteovski, Lawrence, Wu, and
  Neubig}]{viswanathan2023large}
Vijay Viswanathan, Kiril Gashteovski, Carolin~(Haas) Lawrence,
  Tongshuang~Sherry Wu, and Graham Neubig. 2023.
\newblock \href {https://api.semanticscholar.org/CorpusID:259317075} {Large
  language models enable few-shot clustering}.
\newblock \emph{Transactions of the Association for Computational Linguistics},
  12:321--333.

\bibitem[{Wallace et~al.(2020)Wallace, Saha, Soboczenski, and
  Marshall}]{Wallace2020GeneratingN}
Byron~C. Wallace, Sayantani Saha, Frank Soboczenski, and Iain~James Marshall.
  2020.
\newblock \href {https://api.semanticscholar.org/CorpusID:221319573}
  {Generating (factual?) narrative summaries of rcts: Experiments with neural
  multi-document summarization}.
\newblock \emph{AMIA ... Annual Symposium proceedings. AMIA Symposium},
  2021:605--614.

\bibitem[{Wang et~al.(2022)Wang, Li, Pang, He, Li, Tang, and
  Wang}]{wang2022multi}
Pancheng Wang, Shasha Li, Kunyuan Pang, Liangliang He, Dong Li, Jintao Tang,
  and Ting Wang. 2022.
\newblock Multi-document scientific summarization from a knowledge
  graph-centric view.
\newblock \emph{arXiv preprint arXiv:2209.04319}.

\bibitem[{Wang et~al.(2023)Wang, Shang, and Zhong}]{Wang2023GoalDrivenEC}
Zihan Wang, Jingbo Shang, and Ruiqi Zhong. 2023.
\newblock \href {https://api.semanticscholar.org/CorpusID:258841261}
  {Goal-driven explainable clustering via language descriptions}.
\newblock \emph{ArXiv}, abs/2305.13749.

\bibitem[{Wei et~al.(2022{\natexlab{a}})Wei, Wang, Schuurmans, Bosma, hsin Chi,
  Xia, Le, and Zhou}]{Wei2022ChainOT}
Jason Wei, Xuezhi Wang, Dale Schuurmans, Maarten Bosma, Ed~Huai hsin Chi,
  F.~Xia, Quoc Le, and Denny Zhou. 2022{\natexlab{a}}.
\newblock \href {https://api.semanticscholar.org/CorpusID:246411621} {Chain of
  thought prompting elicits reasoning in large language models}.
\newblock \emph{ArXiv}, abs/2201.11903.

\bibitem[{Wei et~al.(2022{\natexlab{b}})Wei, Wang, Schuurmans, Bosma, Xia, Chi,
  Le, Zhou et~al.}]{wei2022chain}
Jason Wei, Xuezhi Wang, Dale Schuurmans, Maarten Bosma, Fei Xia, Ed~Chi, Quoc~V
  Le, Denny Zhou, et~al. 2022{\natexlab{b}}.
\newblock Chain-of-thought prompting elicits reasoning in large language
  models.
\newblock \emph{Advances in Neural Information Processing Systems},
  35:24824--24837.

\bibitem[{Wolf et~al.(2019)Wolf, Debut, Sanh, Chaumond, Delangue, Moi, Cistac,
  Rault, Louf, Funtowicz, and Brew}]{Wolf2019HuggingFacesTS}
Thomas Wolf, Lysandre Debut, Victor Sanh, Julien Chaumond, Clement Delangue,
  Anthony Moi, Pierric Cistac, Tim Rault, R{\'e}mi Louf, Morgan Funtowicz, and
  Jamie Brew. 2019.
\newblock \href {https://api.semanticscholar.org/CorpusID:208117506}
  {Huggingface's transformers: State-of-the-art natural language processing}.
\newblock \emph{ArXiv}, abs/1910.03771.

\bibitem[{Wu et~al.(2021)Wu, Shieh, Hsu, and Chen}]{wu2021towards}
Jia-Yan Wu, Alexander Te-Wei Shieh, Shih-Ju Hsu, and Yun-Nung Chen. 2021.
\newblock Towards generating citation sentences for multiple references with
  intent control.
\newblock \emph{arXiv preprint arXiv:2112.01332}.

\bibitem[{Xing et~al.(2020)Xing, Fan, and Wan}]{xing-etal-2020-automatic}
Xinyu Xing, Xiaosheng Fan, and Xiaojun Wan. 2020.
\newblock \href {https://doi.org/10.18653/v1/2020.acl-main.550} {Automatic
  generation of citation texts in scholarly papers: A pilot study}.
\newblock In \emph{Proceedings of the 58th Annual Meeting of the Association
  for Computational Linguistics}, pages 6181--6190, Online. Association for
  Computational Linguistics.

\bibitem[{Yun et~al.(2023)Yun, Marshall, Trikalinos, and
  Wallace}]{yun-etal-2023-appraising}
Hye Yun, Iain Marshall, Thomas Trikalinos, and Byron Wallace. 2023.
\newblock \href {https://doi.org/10.18653/v1/2023.emnlp-main.626} {Appraising
  the potential uses and harms of {LLM}s for medical systematic reviews}.
\newblock In \emph{Proceedings of the 2023 Conference on Empirical Methods in
  Natural Language Processing}, pages 10122--10139, Singapore. Association for
  Computational Linguistics.

\bibitem[{Zhang et~al.(2023)Zhang, Wang, and Shang}]{Zhang2023ClusterLLMLL}
Yuwei Zhang, Zihan Wang, and Jingbo Shang. 2023.
\newblock \href {https://api.semanticscholar.org/CorpusID:258866119}
  {Clusterllm: Large language models as a guide for text clustering}.
\newblock \emph{ArXiv}, abs/2305.14871.

\bibitem[{Zhu et~al.(2023)Zhu, Feng, Feng, Wu, and
  Qin}]{zhu-etal-2023-hierarchical}
Kun Zhu, Xiaocheng Feng, Xiachong Feng, Yingsheng Wu, and Bing Qin. 2023.
\newblock \href {https://doi.org/10.18653/v1/2023.findings-emnlp.453}
  {Hierarchical catalogue generation for literature review: A benchmark}.
\newblock In \emph{Findings of the Association for Computational Linguistics:
  EMNLP 2023}, pages 6790--6804, Singapore. Association for Computational
  Linguistics.

\end{thebibliography}
\bibliographystyle{acl_natbib}

\appendix

\section{Prompts for Hierarchy Generation Pipeline}
\label{sec:prompt}
\begin{figure*}[]
    \centering
    \includegraphics[width=\textwidth]{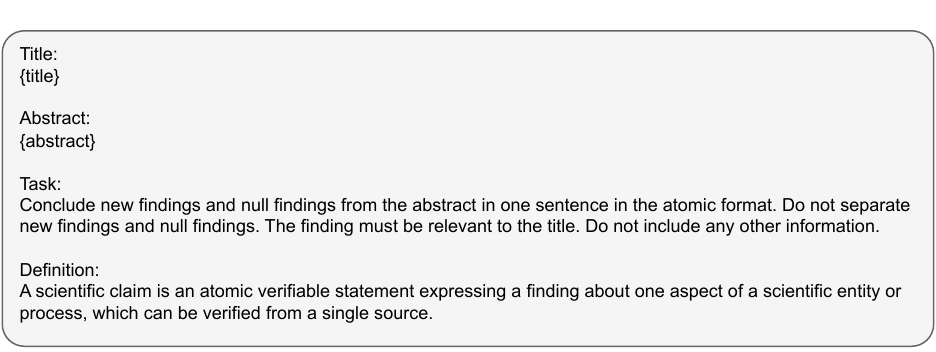}
    \caption{
    Claim generation prompt for GPT-3.5 Turbo.
    }
    \label{fig:claim_prompt}
\end{figure*}

\begin{figure*}[]
    \centering
    \includegraphics[width=\textwidth]{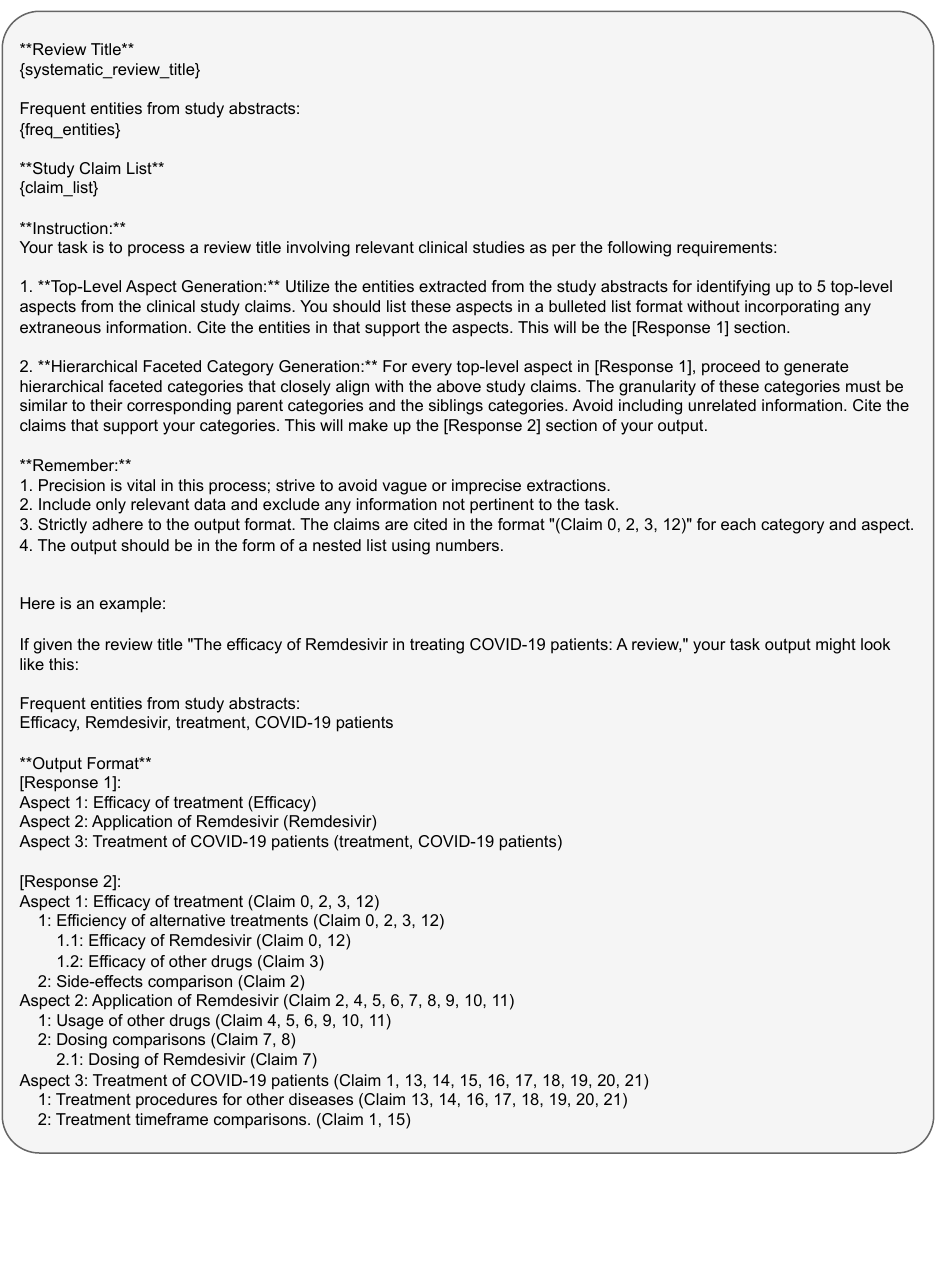}
    \caption{
    Hierarchy proposal module prompt for Claude-2.
    }
    \label{fig:completion_prompt}
\end{figure*}

We present prompts for the hierarchy generation pipeline in \figref{fig:claim_prompt} and \figref{fig:completion_prompt}.

\section{Model Training Details}
\label{appx:model}
\begin{figure*}[]
    \centering
    \includegraphics[width=\textwidth]{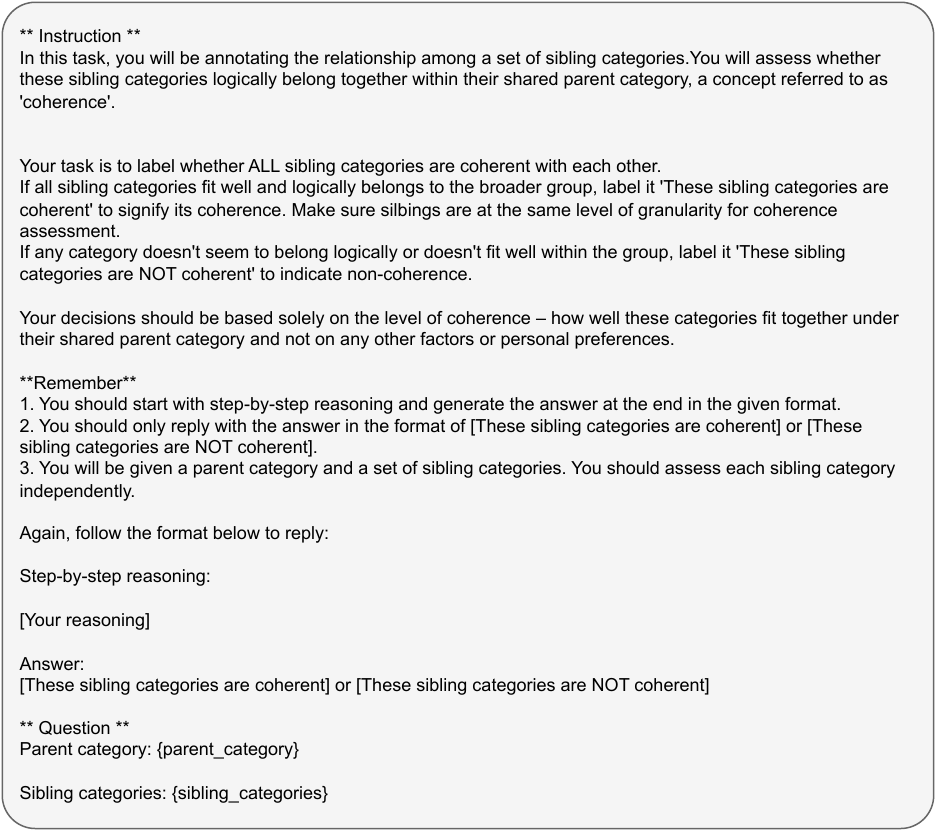}
    \caption{
    Prompt for task 1 sibling coherence for both GPT-3.5 Turbo and GPT-4 Turbo.
    }
    \label{fig:task1_prompt}
\end{figure*}

\begin{figure*}[]
    \centering
    \includegraphics[width=\textwidth]{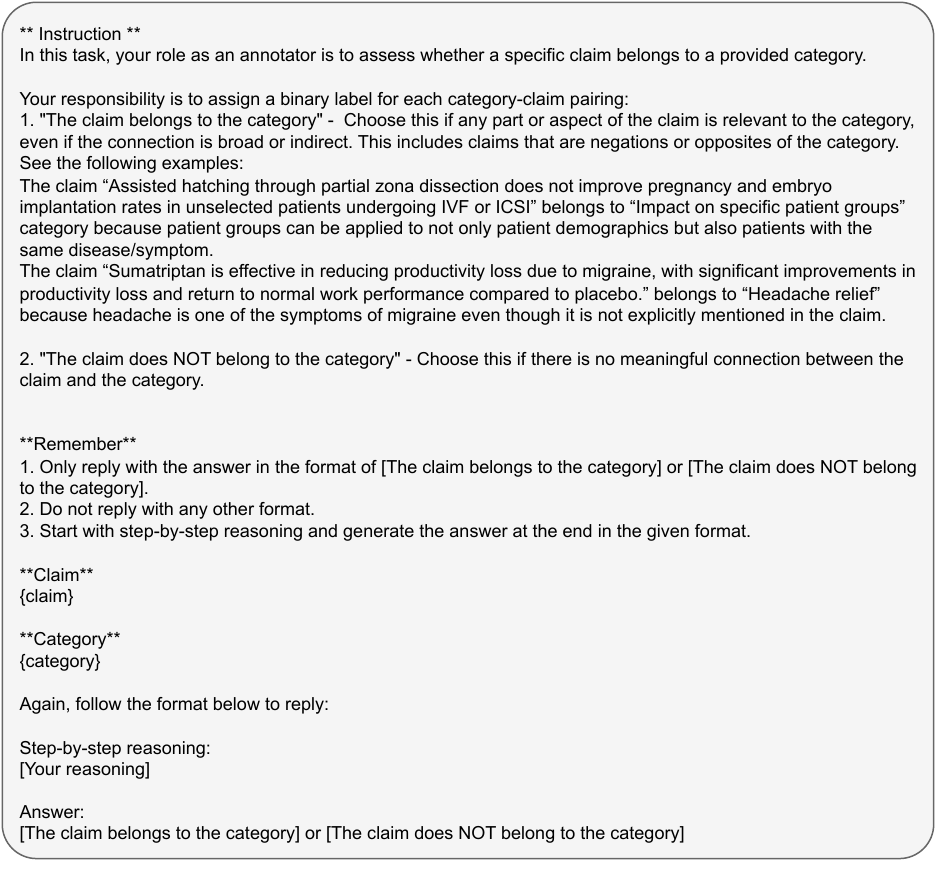}
    \caption{
    Prompt for task 3 claim assignment  for both GPT-3.5 Turbo and GPT-4 Turbo.
    }
    \label{fig:task3_prompt}
\end{figure*}

\paragraph{Flan-T5 fintuning.}
We fine-tuned the \texttt{flan-t5-base} and \texttt{flan-t5-large} models using the Hugggingface library \cite{Wolf2019HuggingFacesTS} with NVIDIA RTX A6000 for both task 1 and task 3. 
For task 1, the learning rate is set to 1e-3 and the batch size is 16. We train the model for up to five epochs.
For task 3, the learning rate is 3e-4 with batch size 16, and the models are trained up to two epochs.
Each epoch takes less than 15 minutes for both model sizes. The numbers reported for each Flan-T5 model come from a single model checkpoint.
\paragraph{GPT-3.5 Turbo and GPT-4 Turbo}
We perform zero-shot CoT prompting for corrector models on tasks 1 and 3 with prompts in \figref{fig:task1_prompt} and \figref{fig:task3_prompt}.

\section{Model Selection for Hierarchy Proposal Module}
\label{sec:model-selection}
We conducted a qualitative evaluation of hierarchies generated by GPT-3.5-Turbo, GPT-4, and \textsc{Claude-2} for 10 sampled research topics. Results showed that GPT-3.5-Turbo does poorly at following instructions and only generates well-formed hierarchies 30\% of the time, while GPT-4 produces valid hierarchies but generates shallow ones with a depth of 1 60\% of the time. In comparison, Claude-2 produces hierarchies with a higher depth (>1) 90\% of the time.

\section{Qualitative Analysis on Generated Claims}
\label{sec:claim-fact}
To better establish the accuracy of our NLI-based verification process, we have conducted an additional qualitative assessment of 100 abstract-claim pairs. We examined 50 pairs that the NLI model marked as “entailed” and 50 non-entailed pairs. Results show that the precision of the NLI model is very high, with 47 out of 50 entailed claims being correct, without hallucinations. Interestingly, we find that 37/50 non-entailed pairs are false negatives, indicating that in many cases, the generated claim is correct even though the NLI model predicts non-entailment. This human evaluation further validates that our claim generation process is high quality.

\section{Qualitative Analysis on Retrieval Quality}
\label{sec:retrieval-quality}
We conducted a brief experiment on 10 samples (sets of related studies present in our dataset) by injecting five irrelevant claims from other study sets per sample. We observed that during hierarchy generation, \textsc{Claude-2} was able to ignore irrelevant claims and generate hierarchies similar to the ones it originally produced (in the non-noisy setting). \textsc{Claude-2} can also differentiate between relevant and irrelevant claims and does not assign noisy claims to any categories in the hierarchy.

\label{sec:appendix}

\end{document}